\documentclass[10pt,letterpaper]{article}
\usepackage[top=0.85in,left=2.75in,footskip=0.75in]{geometry}

\usepackage{amsmath,amssymb}

\usepackage{changepage}

\usepackage{textcomp,marvosym}

\usepackage{cite}

\usepackage{nameref,hyperref}

\usepackage[nopatch=eqnum]{microtype}
\DisableLigatures[f]{encoding = *, family = * }

\usepackage[table]{xcolor}

\usepackage{array}

\newcolumntype{+}{!{\vrule width 2pt}}

\newlength\savedwidth

\raggedright
\usepackage[aboveskip=1pt,labelfont=bf,labelsep=period,justification=raggedright,singlelinecheck=off]{caption}

\makeatletter
\renewcommand{\@biblabel}[1]{\quad#1.}
\makeatother

\usepackage{lastpage,fancyhdr,graphicx}
\usepackage{epstopdf}
\fancyheadoffset[L]{2.25in}
\fancyfootoffset[L]{2.25in}
\usepackage{float}
\usepackage{longtable}
\usepackage{multirow} 
\usepackage{caption}
\begin{document}
\vspace*{0.2in}

\begin{flushleft}
{\Large
\textbf\newline{Machine Learning-Based Model for Postoperative Stroke Prediction in Coronary Artery Disease} 
}
\newline
\\
Haonan Pan\textsuperscript{1},
Shuheng Chen\textsuperscript{1},
Elham Pishgar\textsuperscript{2},
Kamiar Alaei\textsuperscript{3},
Greg Placencia\textsuperscript{4},
Maryam Pishgar\textsuperscript{1*}
\\
\bigskip
\textbf{1} Department of Industrial and Systems Engineering, University of Southern California, 3715 McClintock Ave GER 240, Los Angeles, 90087, California, United States
\\
\textbf{2} Colorectal Research Center, Iran University of Medical Sciences, Tehran Hemat Highway next to Milad Tower, Tehran, 14535, Iran
\\
\textbf{3} Department of Health Science, California State University, Long Beach (CSULB), 1250 Bellflower Blvd, Long Beach, 90840, California, United States
\\
\textbf{4} Department of Industrial and Manufacturing Engineering, California State Polytechnic University, Pomona, 3801 W Temple Ave, Pomona, 91768, California, United States
\\
\bigskip

* pishgar@usc.edu

\end{flushleft}
\section*{Abstract}

\textbf{Background.} Coronary artery disease remains one of the leading causes of mortality globally. Notwithstanding ongoing advancements in revascularization procedures like percutaneous coronary intervention (PCI) and coronary artery bypass grafting (CABG), postoperative stroke persists as a significant and inevitable consequence. This study seeks to create and validate a sophisticated machine learning prediction model to improve the evaluation of postoperative stroke risk in patients having coronary revascularization treatments.

\textbf{Methods.} This research employed data from the MIMIC-IV v3.1 database, consisting of a cohort of 7,023 individuals. Clinical features, laboratory values, and comorbidity information were obtained for study. Variables with over 30\% missing values were removed, and features with a correlation coefficient greater than 0.9 were discarded to mitigate the impact of multicollinearity. The dataset was divided into a training set (70\%) and a test set (30\%). Missing values in the residual dataset were interpolated utilizing the Random Forest algorithm. Numerical values were normalized, whereas categorical variables were transformed by one-hot encoding. Feature selection was subsequently conducted using the least absolute shrinkage and selection operator (LASSO) regularization method, while appropriate model hyperparameters were identified by grid search. Finally, machine learning methods such as Logistic Regression, Extreme Gradient Boosting (XGBoost), Support Vector Machines (SVM), and Categorical Gradient Boosting (CatBoost) were used for predictive modeling, and the effect of each variable on stroke risk was assessed by SHapley Additive Properties (SHAP) analysis. 

\textbf{Results.} The findings demonstrate that the SVM model attained superior predictive performance, evidenced by an area under the curve (AUC) of 0.855 (95\%CI: 0.829-0.878), reflecting substantial enhancements compared to the conventional logistic regression model and the CatBoost model documented in previous studies. SHAP research showed that the Charlson Comorbidity Index (CCI), diabetes, chronic kidney disease (CKD), and heart failure are significant prognostic factors for postoperative stroke.

\textbf{Conclusion.} This study demonstrates that advanced machine learning techniques effectively mitigate overfitting and enhance overall model predictive performance. Furthermore, integrating independent comorbidity factors allows for a more precise prediction of postoperative stroke risk compared to models relying exclusively on the CCI. By incorporating a broader spectrum of clinically relevant variables, the proposed approach offers a more comprehensive and individualized risk assessment, thereby providing a more significant reference for preoperative risk evaluation and tailored intervention.


\section*{Introduction}
Coronary artery disease (CAD), a predominant kind of ischemic heart disease, is the primary cause of mortality globally, resulting in approximately 17.6 million deaths each year from cardiovascular diseases, hence imposing a significant strain on healthcare systems worldwide \cite{bib1,bib2}. For patients with obstructive multivessel disease or intricate coronary architecture, revascularization treatments, including PCI and CABG, are acknowledged as effective therapeutic options \cite{bib3}. Despite continuous improvements in PCI and CABG techniques that have significantly enhanced procedural safety, postoperative stroke remains a serious and unavoidable consequence \cite{bib4}. Postoperative stroke occurs in roughly 1\% to 5\% of patients undergoing cardiac surgery, with in-hospital stroke mortality rates reported to be three to ten times higher than those of community-onset strokes, primarily due to delayed recognition and the increased complexity of medical management \cite{bib5}. Consequently, creating a precise prediction model to identify patients at increased risk of postoperative stroke is crucial. Although machine learning models have shown significant advantages in medical prediction recent years, for example, Sun et al. (2023) employed a logistic regression model to forecast large-artery atherosclerosis (LAA), attaining an AUC of 0.93 in the external validation cohort, signifying robust predictive performance \cite{bib29}; Also using a logistic regression model, de Hond et al. (2022) developed a predictive model for severe asthma exacerbations based on home monitoring data from asthma patients, achieved an AUC of 0.88 in the validation cohort, underscoring the dependability of logistic regression \cite{bib30}; Boros et al. (2025) employed XGBoost and CatBoost machine-learning models to predict in-hospital mortality for patients with acute gastrointestinal bleeding (GIB). In the internal validation set, both models attained an AUC of 0.84 \cite{bib31}; Chen et al. (2025) used the XGBoost algorithm to predict the risk of ICU death in patients with sepsis-associated acute kidney injury, with a model AUROC of 0.878 \cite{bib32}; Huang et al. (2021) utilized SVM model to detect esophageal cancer through breathomics analysis achieving an AUC of 0.89 \cite{bib33}; additionally, Noroozi et al. (2023) employed seven machine learning models (including SVM, Bayes Net, Naïve Bayes, Multivariate Linear Model, Logit Boost, J48, and Random Forest) to predict heart disease using the Cleveland Heart Disease dataset. The best-performing model, SVM, attained an AUC of 0.90 and an accuracy of 85.5\% \cite{bib34}. While these studies have demonstrated the potential of machine learning in medical prediction, the precise prediction of postoperative stroke risk remains a significant challenge. Current models have yet to fully leverage the capabilities of machine learning to accurately identify high-risk individuals undergoing coronary revascularization.

To address this gap, Lin et al. (2024), in their study published in \textit{PLOS ONE}, titled "Machine learning-based models for prediction of the risk of stroke in CAD patients receiving coronary revascularization," investigated the utility of machine learning-based approaches in forecasting postoperative stroke risk. Their study employed CatBoost which acheived an AUC of 0.760 (95\% CI: 0.722–0.798) showing moderate predictive performance in identifying postoperative stroke risk among CAD patients undergoing coronary revascularization\cite{bib7}. However, the study does exhibit certain limitations. First, regarding feature selection, the study employed LASSO to identify and retain 20 features with non-zero coefficients. However, this approach may inadvertently preserve numerous features having low absolute coefficients, leading to excessive noise, complicating the model, and heightening the risk of overfitting \cite{bib21}. Secondly, the research did not address the issue of data imbalance, which is particularly prevalent in medical datasets where case categories frequently exhibit significant disparities, such as the ratio of patients with a disease to healthy controls. Training the model on such unbalanced data may lead to a bias favoring the majority categories, potentially overlooking high-risk patients and consequently undermining the clinical relevance of the predictions \cite{bib35}. Additionally, the absence of hyperparameter optimization techniques, including Grid Search or Bayesian Optimization, during the model training process may lead to a suboptimal level of predictive efficacy in the final model \cite{bib36}. Moreover, in the SHAP analysis of the study, the CCI played a significantly more important role in predicting postoperative stroke compared to other features. Although the CCI, as a composite score, offers a comprehensive evaluation of multiple comorbidities and holds substantial clinical value, its inclusion of many comorbidities is a double-edged sword---while it is highly useful for assessing overall patient health, it may introduce bias when predicting the impact of specific diseases. If the focus shifts to the independent contributions of specific comorbidities, then incorporating certain diseases that have been confirmed to be highly correlated with postoperative stroke---such as diabetes, heart failure, and CKD---separately into the model may further enhance its predictive performance \cite{bib8,bib9}.

This study will address the aforementioned limitations by optimizing feature selection methods to mitigate potential issues arising from low-coefficient noise features. Additionally, the Synthetic Minority Over-sampling Technique (SMOTE) will be incorporated to address data imbalance, thereby reducing bias introduced by the majority class and improving the model’s ability to accurately identify high-risk patients in the minority class. Furthermore, Grid Search will be employed for hyperparameter optimization to further enhance the predictive performance of the model. Recognizing the limitations of excessive reliance on the CCI in existing models, this study will incorporate additional critical risk factors that have been underrepresented in prior models but are strongly associated with postoperative stroke, as identified through a comprehensive literature review and expert consultation. By systematically addressing the shortcomings of Lin et al. (2024), this study aims to develop a more accurate and clinically applicable risk prediction model, providing improved postoperative stroke risk assessment and clinical decision support for CAD patients undergoing coronary revascularization.

\section*{Methods}

\subsection*{Data Source}
The data for this research were obtained from the MIMIC-IV v3.1 (Medical Information Mart for Intensive Care IV) database. MIMIC-IV is a publicly accessible database that includes comprehensive data on patients hospitalized in a tertiary academic medical center in Boston, MA, USA. The database records the length of each patient's stay, laboratory tests, prescription protocols, vital signs, and additional specific information during their ICU admission, thereby providing a valuable resource for clinical research and machine learning in critical care \cite{bib10}.

\subsection*{Study Population}
\begin{figure}[H]
    \centering
    \includegraphics[width=1\linewidth]{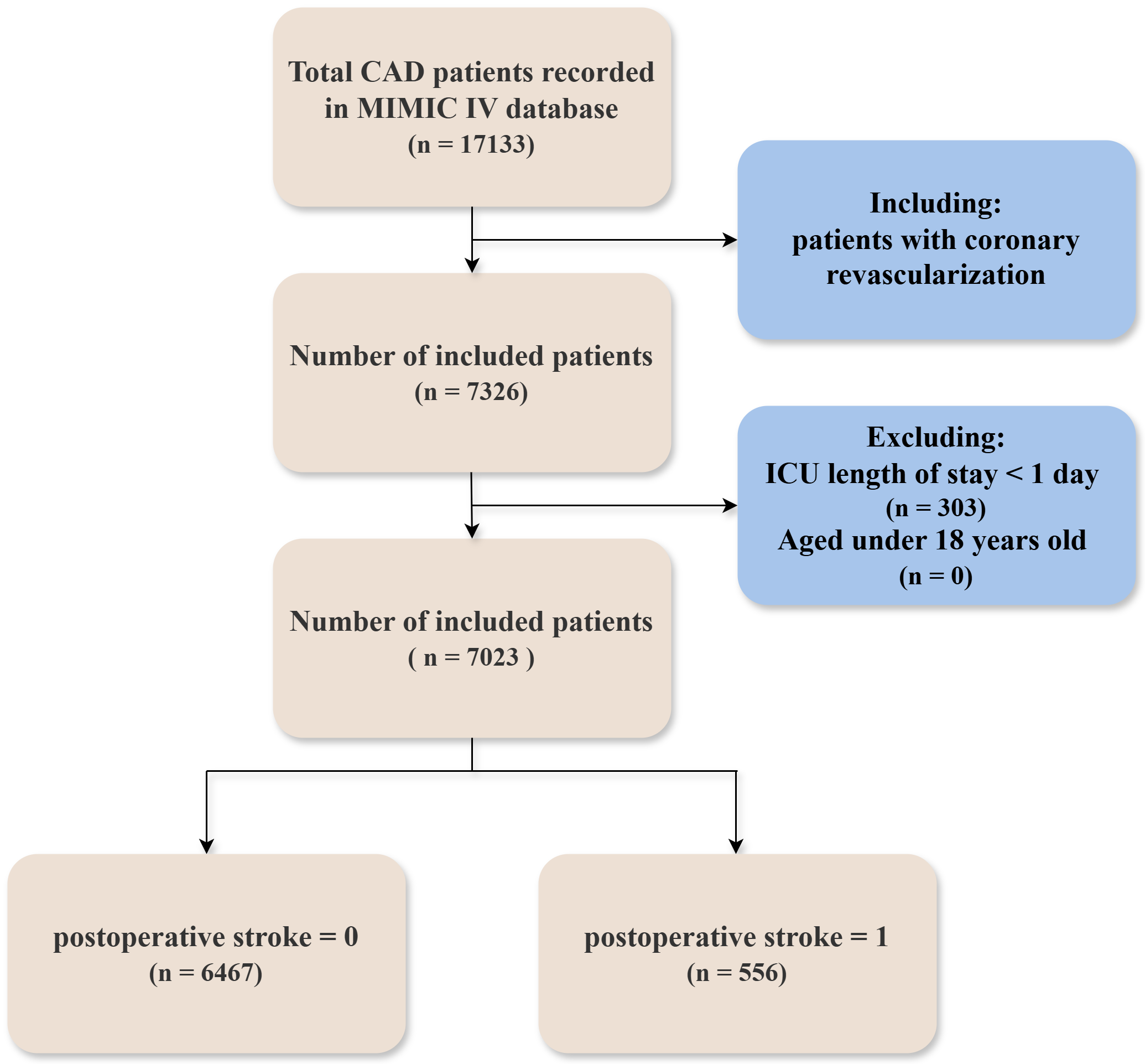}
    \caption{\bf Patient Selection Flowchart for Coronary Revascularization in the MIMIC-IV Database.}
    \label{fig1}
\end{figure}

A comprehensive analysis revealed that 17,133 individuals diagnosed with CAD were documented within the MIMIC-IV database. A total of 7,326 patients received coronary revascularization, which encompassed PCI, CABG, or a combination of these procedures. In order to enhance the study cohort, we implemented exclusion criteria, eliminating 303 patients who had an ICU length of stay of less than one day, while no patients under the age of 18 were found to warrant exclusion. Subsequent to these exclusions, the final analysis encompassed a total of 7,023 patients. The patients were subsequently classified according to the incidence of postoperative stroke, revealing that 6,467 patients (92.1\%) did not experience such an event, whereas 556 patients (7.9\%) were identified as having postoperative stroke. The process of selecting patients is depicted in (Fig~\ref{fig1}), illustrating the criteria for inclusion and exclusion that culminate in the final study cohort.

\subsection*{Data Collection}

The date collection process was conducted by incorporating a comprehensive set of clinically recognized stroke risk factors, specific variables were determined to be included for model training. Throughout this process, we consulted the most recent evidence-based studies and expert insights to guarantee that the chosen features thoroughly and precisely represent the patients' stroke risk.

The variables considered in the final dataset included demographic factors such as age, gender, ethnicity, insurance type (Medicaid, Medicare, or others), and marital status; clinical factors such as first care unit (CCU, CVICU, or others), family history of stroke, personal history of stroke, treatments (CABG, PCI, or both), thrombolysis, use of antiplatelets, beta-blockers, calcium channel blockers, ventilation, vasopressors, and other medications such as angiotensin-converting enzyme inhibitors (ACEI), antibiotics, angiotensin II receptor blockers (ARB), Non-steroidal anti-inflammatory drug (NSAID); neurological and comorbidity assessments including Glasgow Coma Scale (GCS), CCI, SOFA score, and conditions such as hypertension, diabetes, CKD, hyperlipidemia, obesity, heart failure, and Peripheral Vascular Disease (PVD); vital signs including weight, heart rate, systolic blood pressure (SBP), diastolic blood pressure (DBP), respiratory rate, temperature, and oxygen saturation (SpO\textsubscript{2}); and laboratory measurements comprising a comprehensive blood test (CBC) with hematocrit, hemoglobin, mean corpuscular hemoglobin (MCH), mean corpuscular hemoglobin concentration (MCHC), mean corpuscular volume (MCV), platelet count, red blood cell count (RBC), red cell distribution width (RDW), red cell distribution width standard deviation (RDW-SD), white blood cell count (WBC), creatinine, international normalized ratio (INR), prothrombin time (PT), partial thromboplastin time (PTT), blood urea nitrogen (BUN), glucose, calcium, sodium, chloride, bicarbonate, and lactate.

In order to improve the clinical applicability of the model, we meticulously regulated the timeframe for data collection, guaranteeing that all variables were obtained within 24 hours of ICU admission.

\subsection*{Data Preprocessing}
The dataset initially comprised a total of 103 features. By setting a threshold of 0.3 and eliminating all features with over 30\% missing data, 76 features remained. Prior to further analysis, the data were partitioned into a training set (n=4916) and a testing set (n=2107) in a 7:3 ratio to avert information leakage \cite{bib17}. All absent categorical data were replaced with ‘Unknown’, and a Random Forest Imputer was used to address missing data in the training set; the same imputation strategy was subsequently applied to the testing set. A standard scaler was used to preprocess all numerical data in the training set and then applied to the testing set. After applying one-hot encoding to the categorical data, a total of 88 features were preserved.

\subsection*{Feature Selection}
\begin{figure}[H]
    \centering
    \includegraphics[width=1\linewidth]{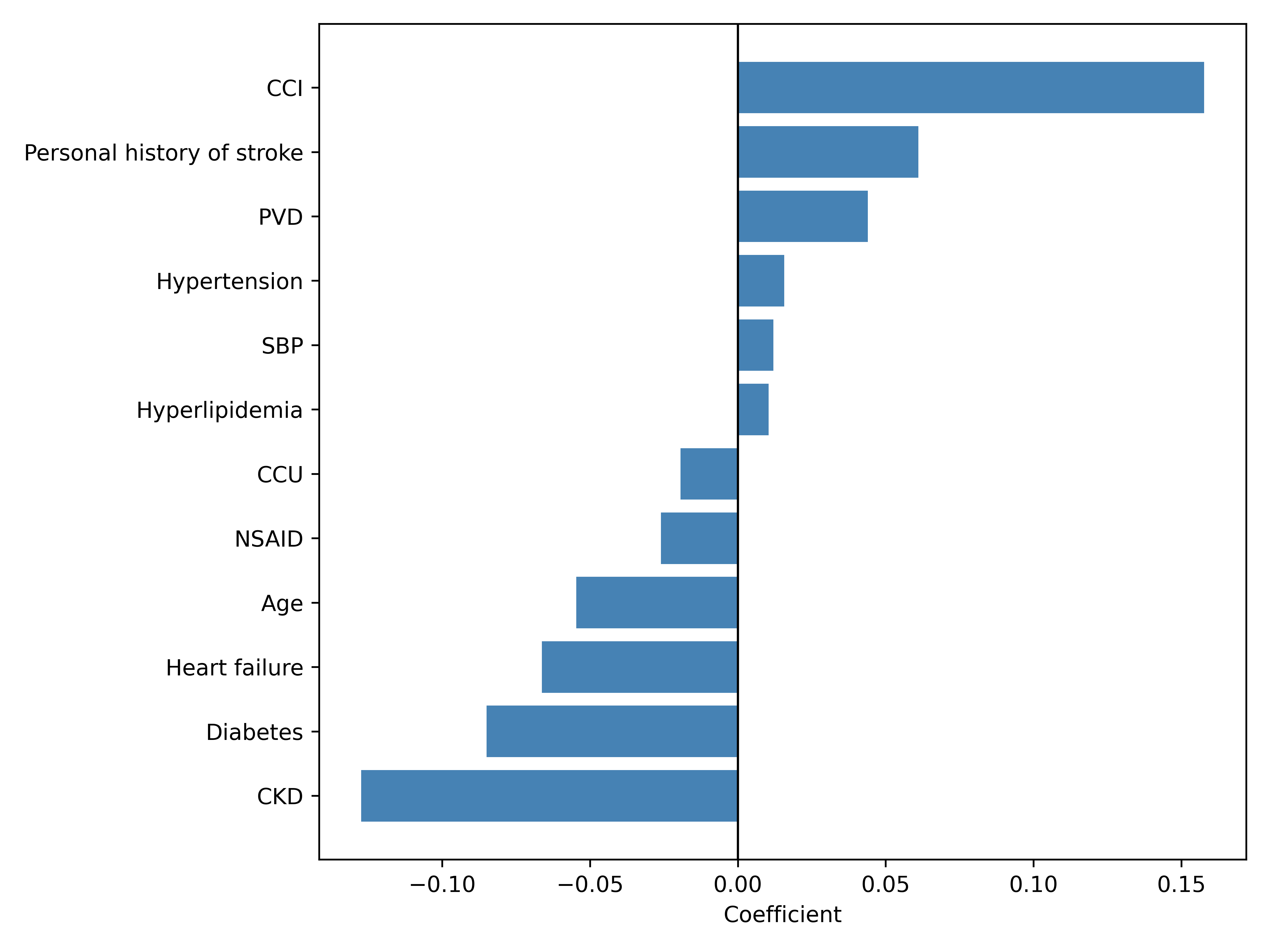}
    \caption{\bf Features with Absolute Coefficient Greater Than 0.01.}
    \label{fig2}
\end{figure}

\begin{figure}[H]
    \centering
    \includegraphics[width=1\linewidth]{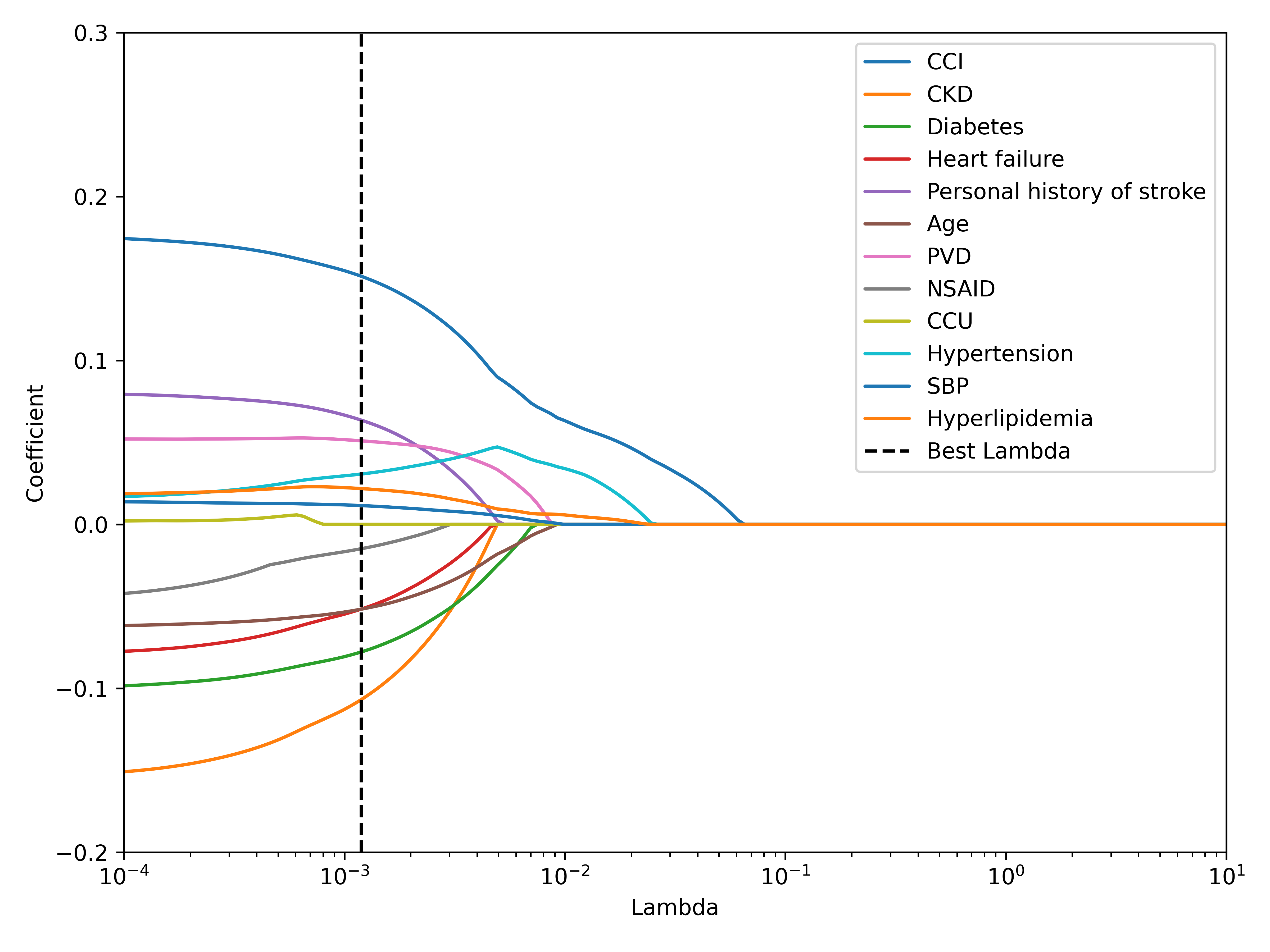}
    \caption{\bf Regularization Path of LASSO Regression.}
    \label{fig3}
\end{figure}

Following data preprocessing, the feature space was simplified by removing features with a correlation greater than 90\%, leaving a total of 79 features. Thereafter, a 10-fold cross-validation LASSO regression was employed on the remaining 79 features (random state = 42) in order to identify the independent variables with the greatest predictive capacity, thereby minimizing noise interference while optimizing the preservation of essential information. Parameter tuning yielded an optimal regularization parameter ($\lambda = 0.0011908$). Features with absolute coefficients exceeding 0.01 were then retained, resulting in 12 selected variables (Fig~\ref{fig2}): CCI, CKD, Diabetes, Heart Failure, Personal History of Stroke, Age, PVD, NSAID, First Care Unit (CCU), Hypertension, SBP, and Hyperlipidemia. Fig~\ref{fig3} displays the corresponding LASSO path plot.


\subsection*{Model Development}

In this study, we collected comprehensive and multi-dimensional clinical information from the MIMIC-IV database and followed the aforementioned data cleaning and feature selection procedures.  Since the target variable of this study, postoperative stroke, is highly imbalanced within the patient population (see Fig~\ref{fig1}), we employed the SMOTE (Synthetic Minority Class Oversampling) on the training set to generate additional "minority class" samples that resemble the real distribution. This approach mitigating the model's bias in  identifying minority classes \cite{bib20}. It is worth noting that SMOTE was applied only to the training set to prevent potential data leakage during model evaluation; the test set retained its original class distribution to ensure a reliable evaluation of the model's generalization capability in a real clinical setting.

Following the implementation of data balancing, the Grid Search and Cross-Validation methods were employed to systematically explore the optimal hyperparameters of each model. Through the iteration of numerous candidate parameter combinations, the ideal configuration was identified (see Table~\ref{table2}). This configuration achieved a balance between the model's capacity to learn complex clinical features and its generalization performance.

The proposed model of this study is SVM, which can effectively handle complex nonlinear relationships within the data by using kernel functions to map the data into a higher dimensional space. This characteristic enables SVM to often exhibit superior generalization capabilities in intricate classification applications. At the same time, with proper parameter tuning and regularization constraints, SVM can proficiently mitigate the issue of overfitting. Furthermore, by identifying the maximum-margin hyperplane to differentiate various classes, SVM establishes distinct and unambiguous decision boundaries, while also demonstrating enhanced robustness against noise \cite{bib22}.

To further validate the performance of SVM, we selected several common and representative machine learning algorithms - including XGBoost, CatBoost, and logistic regression - as baseline models for comparison, all of which have been extensively applied and validated when dealing with classification tasks in the medical field. All models are strictly trained according to the same train-test split,  and evaluated using multiple performance metrics such as AUC, sensitivity, specificity, and accuracy, thereby ensuring the objectivity of the results and the equity of the comparison.

\begin{table}[H]
\begin{adjustwidth}{-2.25in}{0in}
\centering
\caption{{\bf Optimal Hyperparameters for Machine Learning Models}}
\label{table2}
\begin{tabular}{l|l|l|l}
\hline
\textbf{XGBoost} & \textbf{LogisticRegression} & \textbf{SVM} & \textbf{CatBoost} \\ \hline
\parbox[t]{4cm}{%
eval\_metric = logloss\\
learning\_rate = 0.05\\
max\_depth = 3\\
n\_estimators = 200\\
random\_state = 42
} & 
\parbox[t]{4cm}{%
C = 0.1\\
max\_iter = 1000\\
n\_jobs = -1\\
penalty = l2\\
solver = liblinear\\
warm\_start = false\\
random\_state = 42\\
 
} & 
\parbox[t]{4cm}{%
C = 1\\
gamma = 0.01\\
kernel = rbf\\
max\_iter = -1\\
probability = true\\
random\_state = 42
} & 
\parbox[t]{4cm}{%
iterations = 500\\
learning\_rate = 0.05\\
depth = 3\\
verbose = false\\
random\_state = 42
} \\ \hline
\end{tabular}
\end{adjustwidth}
\end{table}

\subsection*{Statistical Analyses}
All data were analyzed using Python and its data analysis libraries. The dataset was divided into three types of variables: numerical, binary, and categorical. Continuous variables are presented as mean $\pm$ standard deviation and compared between groups using an independent samples t-test, with the p-value calculated to assess statistical significance. Categorical variables were converted into binary variables using one-hot encoding and are presented, along with the original binary variables, as frequencies and percentages. A chi-square test ($\chi^2$ test) was used for inter-group comparisons, with the p-value calculated to assess the statistical differences in variable distribution. The baseline characteristics are organized and presented in Table~\ref{table1} to provide basic statistical information regarding the study subjects and ensure group comparability.

\section*{Results}

\subsection*{Descriptive Statistics}
Table~\ref{table1} outlines the baseline characteristics of the study population, categorizing them according to the occurrence of postoperative stroke. Out of a total of 7,023 patients, the training set contains 388 (8\%) stroke patients and 4,528 (92\%) non-stroke patients. The testing set includes 168 (8\%) stroke patients and 1,939 (92\%) non-stroke patients.

Regarding gender distribution, although there are generally more male patients, the postoperative stroke incidence is higher in female patients (approximately 7\% for males and 10\% for females, \textit{p} = 0.002 for the training set, \textit{p} = 0.016 for the test set). Age is a substantial risk factor; in both cohorts, the average age of stroke patients is much higher than that of non-stroke patients (training set: 72.0 $\pm$ 9.9 years vs. 68.0 $\pm$ 10.8 years, \textit{p} $<$ 0.001; test set: 71.0 $\pm$ 9.8 years vs. 67.9 $\pm$ 10.5 years, \textit{p} $<$ 0.001). SBP also showed significant differences between the groups, with stroke patients having higher values (training set: 113.8 $\pm$ 10.6 mmHg vs. 112.0 $\pm$ 9.0 mmHg, \textit{p} = 0.002; test set: 115.6 $\pm$ 11.0 mmHg vs. 112.4 $\pm$ 9.3 mmHg, \textit{p} $<$ 0.001).

Patients who suffer postoperative strokes are more likely to have Medicare insurance (\textit{p} $<$ 0.001), although this correlation does not imply that Medicare insurance is a causative factor for strokes. Medicare predominantly serves the older demographic, and as age increases and the likelihood of multiple chronic diseases rises, its higher stroke rate is more likely to reflect the impact of age and overall health condition, rather than a direct causal effect of the insurance type on stroke risk. In relation to treatment plans, whether patients received CABG exclusively, PCI exclusively, or both concurrently, the occurrence of postoperative stroke showed no statistically significant difference in both the training set (\textit{p} = 0.585) and the test set (\textit{p} = 0.198).

From the medical history perspective, compared to a family history of stroke (training set \textit{p} = 0.216, test set \textit{p} = 0.18), the patient's previous history of stroke is more strongly correlated with postoperative stroke (\textit{p} $<$ 0.001). Patients undergoing antiplatelet therapy had a diminished incidence of postoperative stroke (\textit{p} $<$ 0.001), whereas those administered NSAIDs demonstrated a notable reduction in postoperative stroke risk (training set \textit{p} $<$ 0.001, test set \textit{p} = 0.014), suggesting that these two types of drugs may help reduce the risk of postoperative stroke.

Patients with CKD have a significantly increased risk of stroke post-surgery (training set \textit{p} = 0.013, test set \textit{p} = 0.027), underscoring a strong relationship between CKD and postoperative stroke. Concurrently, individuals with PVD showed a significantly greater incidence of postoperative stroke compared to those without the condition (\textit{p} $<$ 0.001). In patients with heart failure, the \textit{p}-value in the training set was 0.062 (marginally above the significance level), but in the test set, the \textit{p}-value was 0.042 (statistically significant), indicating that the relationship between heart failure and postoperative stroke may be more complex. Diabetes, hypertension, and hyperlipidemia had \textit{p}-values greater than 0.05 in both the training and test sets, indicating no significant correlation with stroke. Finally, the CCI score for patients who experienced a stroke post-surgery was significantly higher than that of patients who did not have a stroke (training set: 6.5 $\pm$ 2.2 vs. 4.4 $\pm$ 2.3, \textit{p} $<$ 0.001; test set: 6.6 $\pm$ 2.4 vs. 4.4 $\pm$ 2.3, \textit{p} $<$ 0.001), further indicating that the greater the number of comorbidities, the higher the risk of postoperative stroke.

\begin{longtable}{l|c|c|c|c|c|c|c|c}
\caption{\bf Comparison of Baseline Characteristics Between Patients with and without Postoperative Stroke}
\label{table1}\\
\hline
\multirow{2}{*}{\bf Variable} & \multicolumn{4}{c|}{\bf Training set} & \multicolumn{4}{|c}{\bf Testing set} \\ \cline{2-9} 

 & Yes (8\%) & No (92\%) & $\chi^2$/t & p-value & Yes (8\%) & No (92\%) & $\chi^2$/t & p-value \\ \hline
\endfirsthead
\multicolumn{0}{c}{{\bfseries Table \thetable\ Continued}}\\[0.5ex]
\hline
\multirow{2}{*}{\bf Variable} & \multicolumn{4}{c|}{\bf Training set} & \multicolumn{4}{|c}{\bf Testing set} \\ \cline{2-9} 
 & Yes (8\%) & No (92\%) & $\chi^2$/t & p-value & Yes (8\%) & No (92\%) & $\chi^2$/t & p-value \\ \hline
\endhead
\textbf{Gender\footnotemark[1]} &  &  & 9.448 & 0.002 &  &  & 5.756 & 0.016 \\ \hline
\quad Male (\%) & 273 (70) & 3503 (77) &  &  & 113 (67) & 1472 (76) &  &  \\ \hline
\quad Female (\%) & 115 (30) & 1025 (23) &  &  & 55 (33)  & 467 (24)  &  &  \\ \hline
\textbf{Ethnicity} &  &  & 8.777 & 0.067 &  &  & 0.902 & 0.924 \\ \hline
\quad White (\%) & 290 (75) & 3095 (68) &  &  & 123 (73) & 1389 (72) &  &  \\ \hline
\quad Black (\%) & 14 (4)   & 159 (4)   &  &  & 5 (3)    & 86 (4)    &  &  \\ \hline
\quad Asian (\%) & 6 (2)    & 114 (3)   &  &  & 3 (2)    & 36 (2)    &  &  \\ \hline
\quad Hispanic (\%) & 6 (2)    & 136 (3)   &  &  & 5 (3)    & 51 (3)    &  &  \\ \hline
\quad Others (\%) & 72 (19)  & 1024 (23) &  &  & 32 (19)  & 377 (19)  &  &  \\ \hline
\textbf{Age (years)} & 72.0$\pm$9.9 & 68.0$\pm$10.8 & -7.49 & $<$0.001 & 71.9$\pm$8.5  & 67.9$\pm$10.7 & -5.69 & $<$0.001 \\ \hline
\textbf{SBP (mmHg)} & 113.8$\pm$10.6 & 112.0$\pm$9.0 & -3.17 & 0.002 & 115.6$\pm$11.0 & 112.4$\pm$9.3 & -3.59 & $<$0.001 \\ \hline
\textbf{Insurance Type} &  &  & 30.431 & $<$0.001 &  &  & 22.276 & $<$0.001 \\ \hline
\quad Medicare (\%) & 270 (70) & 2572 (57) &  &  & 125 (74) & 1085 (56) &  &  \\ \hline
\quad Medicaid (\%) & 28 (7)   & 343 (8)   &  &  & 9 (5)    & 162 (8)   &  &  \\ \hline
\quad Others (\%) & 84 (22)  & 1575 (35) &  &  & 32 (19)  & 673 (35)  &  &  \\ \hline
\quad Unknown (\%) & 6 (2)    & 38 (1)    &  &  & 2 (1)    & 19 (1)    &  &  \\ \hline
\textbf{Treatments} &  &  & 1.071 & 0.585 &  &  & 3.243 & 0.198 \\ \hline
\quad CABG (\%) & 334 (86) & 3866 (85) &  &  & 151 (90) & 1643 (85) &  &  \\ \hline
\quad PCI (\%) & 48 (12)  & 613 (14)  &  &  & 16 (10)  & 277 (14)  &  &  \\ \hline
\quad Both (\%) & 6 (2)    & 49 (1)    &  &  & 1 (1)    & 19 (1)    &  &  \\ \hline
\textbf{First Care Unit} &  &  & 10.653 & 0.005 &  &  & 1.042 & 0.594 \\ \hline
\quad CCU (\%) & 48 (12\%)  & 644 (14\%)  &  &  & 21 (12\%)  & 286 (15\%)  &  &  \\ \hline
\quad CVICU (\%) & 329 (85\%) & 3837 (85\%) &  &  & 145 (86\%) & 1639 (85\%) &  &  \\ \hline
\quad Others (\%) & 11 (3\%)   & 47 (1\%)    &  &  & 2 (1\%)    & 14 (1\%)    &  &  \\ \hline
\textbf{Personal History Of Stroke} &  &  & 45.259 & $<$0.001 &  &  & 31.682 & $<$0.001 \\ \hline
\quad Yes (\%) & 55 (14.2) & 248 (5.5) &  &  & 27 (16.1) & 98 (5.1) &  &  \\ \hline
\quad No (\%)  & 333 (85.8) & 4280 (94.5) &  &  & 141 (83.9) & 1841 (94.9) &  &  \\ \hline
\textbf{Family History Of Stroke} &  &  & 1.529 & 0.216 &  &  & 1.796 & 0.180 \\ \hline
\quad Yes (\%) & 42 (10.8) & 596 (13.2) &  &  & 26 (15.5) & 226 (11.7) &  &  \\ \hline
\quad No (\%)  & 346 (89.2) & 3932 (86.8) &  &  & 142 (84.5) & 1713 (88.3) &  &  \\ \hline
\textbf{Anti-Platelet} &  &  & 16.956 & $<$0.001 &  &  & 11.431 & 0.001 \\ \hline
\quad Yes (\%) & 256 (66.0) & 3422 (75.6) &  &  & 107 (63.7) & 1470 (75.8) &  &  \\ \hline
\quad No (\%)  & 132 (34.0) & 1106 (24.4) &  &  & 61 (36.3) & 469 (24.2) &  &  \\ \hline
\textbf{NSAID} &  &  & 21.171 & $<$0.001 &  &  & 6.042 & 0.014 \\ \hline
\quad Yes (\%) & 234 (60.3) & 3239 (71.5) &  &  & 105 (62.5) & 1392 (71.8) &  &  \\ \hline
\quad No (\%)  & 154 (39.7) & 1289 (28.5) &  &  & 63 (37.5)  & 547 (28.2)  &  &  \\ \hline
\textbf{Diabetes} &  &  & 0.472 & 0.492 &  &  & 1.057 & 0.304 \\ \hline
\quad Yes (\%) & 171 (44.1) & 1908 (42.1) &  &  & 76 (45.2) & 792 (40.8) &  &  \\ \hline
\quad No (\%)  & 217 (55.9) & 2620 (57.9) &  &  & 92 (54.8) & 1147 (59.2) &  &  \\ \hline
\textbf{CKD} &  &  & 6.116 & 0.013 &  &  & 4.893 & 0.027 \\ \hline
\quad Yes (\%) & 91 (23.5)  & 825 (18.2)  &  &  & 43 (25.6)  & 355 (18.3)  &  &  \\ \hline
\quad No (\%)  & 297 (76.5)  & 3703 (81.8) &  &  & 125 (74.4) & 1584 (81.7) &  &  \\ \hline
\textbf{Heart Failure} &  &  & 3.479 & 0.062 &  &  & 4.129 & 0.042 \\ \hline
\quad Yes (\%) & 117 (30.2) & 1163 (25.7) &  &  & 57 (33.9)  & 511 (26.4)  &  &  \\ \hline
\quad No (\%)  & 271 (69.8) & 3365 (74.3) &  &  & 111 (66.1) & 1428 (73.6) &  &  \\ \hline
\textbf{Hypertension} &  &  & 0.454 & 0.501 &  &  & 2.108 & 0.147 \\ \hline
\quad Yes (\%) & 238 (61.3) & 2692 (59.5) &  &  & 89 (53.0)  & 1145 (59.1) &  &  \\ \hline
\quad No (\%)  & 150 (38.7) & 1836 (40.5) &  &  & 79 (47.0)  & 794 (40.9)  &  &  \\ \hline
\textbf{Hyperlipidemia} &  &  & 1.909 & 0.167 &  &  & 0.567 & 0.452 \\ \hline
\quad Yes (\%) & 314 (80.9) & 3521 (77.8) &  &  & 133 (79.2) & 1479 (76.3) &  &  \\ \hline
\quad No (\%)  & 74 (19.1)  & 1007 (22.2) &  &  & 35 (20.8)  & 460 (23.7)  &  &  \\ \hline
\textbf{PVD} &  &  & 130.434 & $<$0.001 &  &  & 38.935 & $<$0.001 \\ \hline
\quad Yes (\%) & 94 (24.2)  & 326 (7.2)   &  &  & 36 (21.4)  & 140 (7.2)   &  &  \\ \hline
\quad No (\%)  & 294 (75.8) & 4202 (92.8) &  &  & 132 (78.6) & 1799 (92.8) &  &  \\ \hline
\textbf{CCI} & 6.5$\pm$2.2 & 4.4$\pm$2.3 & -17.23 & $<$0.001 & 6.6$\pm$2.4 & 4.4$\pm$2.3 & -11.19 & $<$0.001 \\ \hline
\end{longtable}
\begin{adjustwidth}{-2.25in}{0in}
\begin{flushleft}
Notes: CKD, Chronic Kidney Disease; SBP, Systolic Blood Pressure; CABG, Coronary Artery Bypass Grafting; PCI, Percutaneous Coronary Intervention; CCI, Charlson Comorbidity Index; NSAID, Non-steroidal anti-inflammatory drug; PVD, Peripheral Vascular Disease.
\end{flushleft}
\end{adjustwidth}

\subsection*{Model Performance Evaluation}
\begin{figure}[H]
    \centering
    \includegraphics[width=1\linewidth]{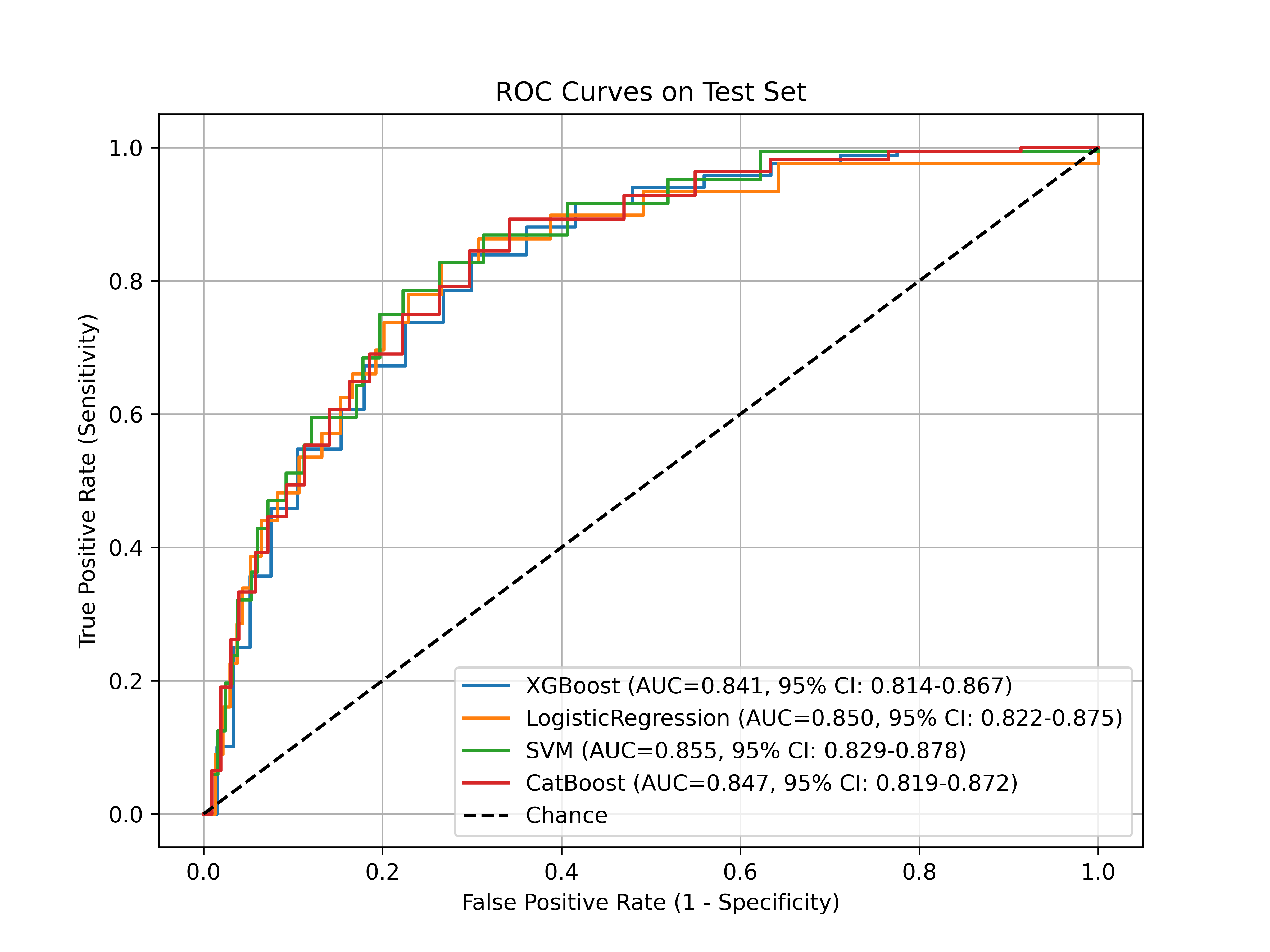}
    \caption{\bf ROC Curves Comparing XGBoost, Logistic Regression, SVM, and CatBoost on the Test Set.}
    \label{fig4}
\end{figure}

Table~\ref{table3} summarizes the predictive performance for postoperative stroke. This study evaluated the performance of four models—XGBoost, Logistic Regression, SVM, and CatBoost—in predicting the risk of postoperative stroke, using metrics such as AUC, sensitivity, specificity, and accuracy. The AUC for each model was as follows (Figure~\ref{fig4}): XGBoost: 0.841 (95\% CI: 0.814 - 0.867); Logistic Regression: AUC: 0.850 (95\% CI: 0.822 - 0.875); SVM: 0.855 (95\% CI: 0.829 - 0.878); CatBoost: 0.847 (95\% CI: 0.819 - 0.872). Although the overall performance of the models was similar, the SVM model demonstrated the best prediction ability on the test set, attaining the highest AUC of 0.86, with balanced performance in sensitivity (0.720), specificity (0.817), and accuracy 0.809 (95\% CI: 0.792 - 0.826). In comparison to traditional Logistic Regression, SVM demonstrated marginal superiority in sensitivity and overall discriminative ability (AUC).

\begin{table}[H]
\begin{adjustwidth}{-1.25in}{0in}
\caption{Model Performance Metrics on Testing Dataset}
\label{table3}
\begin{tabular}{l|l|c|c|c}
\hline
Model  & AUC (95\%CI) & Sensitivity & Specificity & Accuracy (95\%CI) \\
\hline
XGBoost             & 0.841 (0.814 - 0.867) & 0.500 & 0.908 &  0.875 (0.861 - 0.889)
 \\ \hline
LogisticRegression  & 0.850 (0.822 - 0.875) & 0.685 & 0.817 & 0.807 (0.790 - 0.823) \\ \hline
SVM                 & 0.855 (0.829 - 0.878) & 0.720 & 0.817 & 0.809 (0.792 - 0.826) \\ \hline
CatBoost            & 0.847 (0.819 - 0.872) & 0.429 & 0.932 & 0.892 (0.878 - 0.905) \\ \hline
\hline
\end{tabular}
\end{adjustwidth}
\end{table}

\subsection*{SHAP Feature Importance Analysis}

\begin{figure}[H]
    \centering
    \includegraphics[width=1\linewidth]{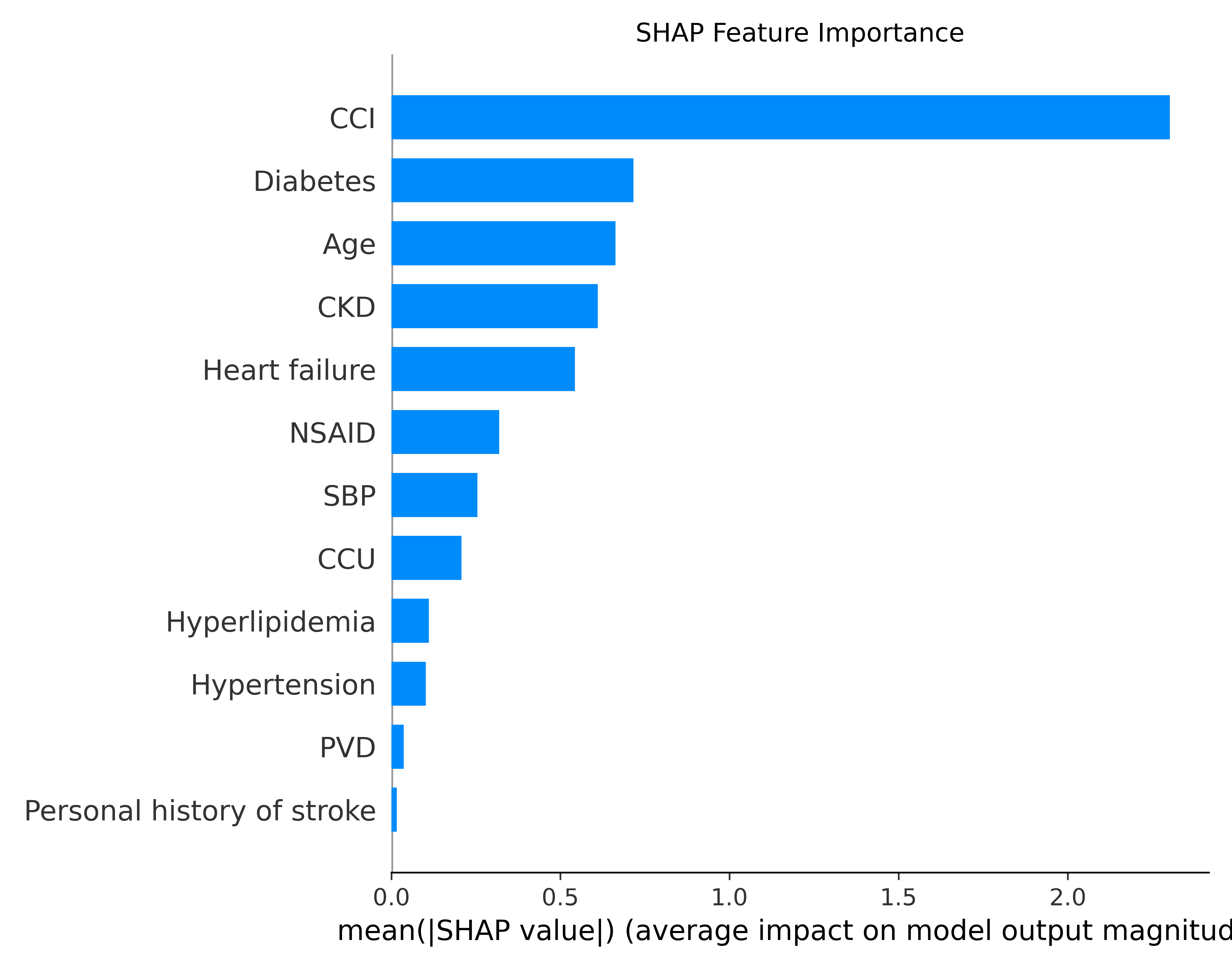}
    \caption{SHAP Bar Summary Plot Showing Feature Importance Based on Absolute Shapley Values.}
    \label{fig5}
\end{figure}

\begin{figure}[H]
    \centering
    \includegraphics[width=1\linewidth]{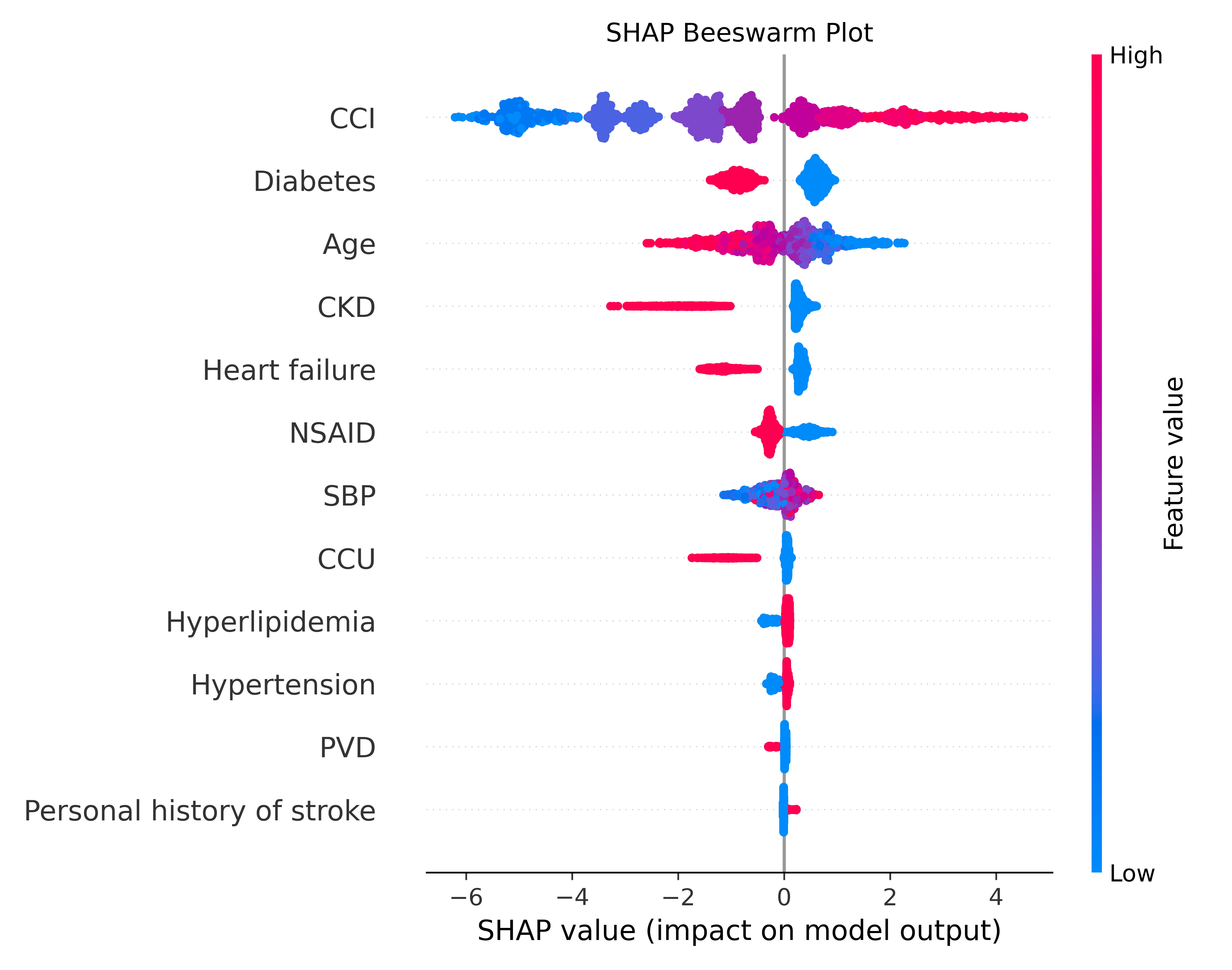}
    \caption{SHAP Beeswarm Plot Illustrating the Impact of Features on Model Predictions.}
    \label{fig6}
\end{figure}

The SHapley Additive exPlanations (SHAP) analysis results (Fig~\ref{fig5} and \ref{fig6}) illustrated the importance and directional impact of various features on predicting postoperative stroke. The SHAP Bar Plot (Fig~\ref{fig5}) indicates that the CCI has the highest predictive contribution, followed by diabetes, age, CKD, and heart failure, highlighting the significant impact of these variables on postoperative stroke risk prediction.

Moreover, the SHAP Beeswarm Plot (Fig~\ref{fig6}) offers further insight into the impact of varying feature values (red indicating higher values, blue indicating lower values) on the model's output. A high SHAP score indicates an elevated stroke risk, whereas a negative SHAP value signifies a reduced stroke risk. As expected, elevated CCI scores (red) correlated with heightened predicted risk (positive SHAP values), confirming its significance as an essential measure of comorbidity burden.

Nonetheless, an unusual pattern was observed with diabetes, age, CKD, and heart failure, wherein elevated feature values were associated with negative SHAP values, indicating that advanced age and the presence of these comorbidities correlate with a diminished anticipated risk of postoperative stroke. This phenomenon likely stems from the intricate composition of the CCI, which effectively encapsulates the overall burden of multimorbidity.

\begin{figure}[H]
    \centering
    \includegraphics[width=1\linewidth]{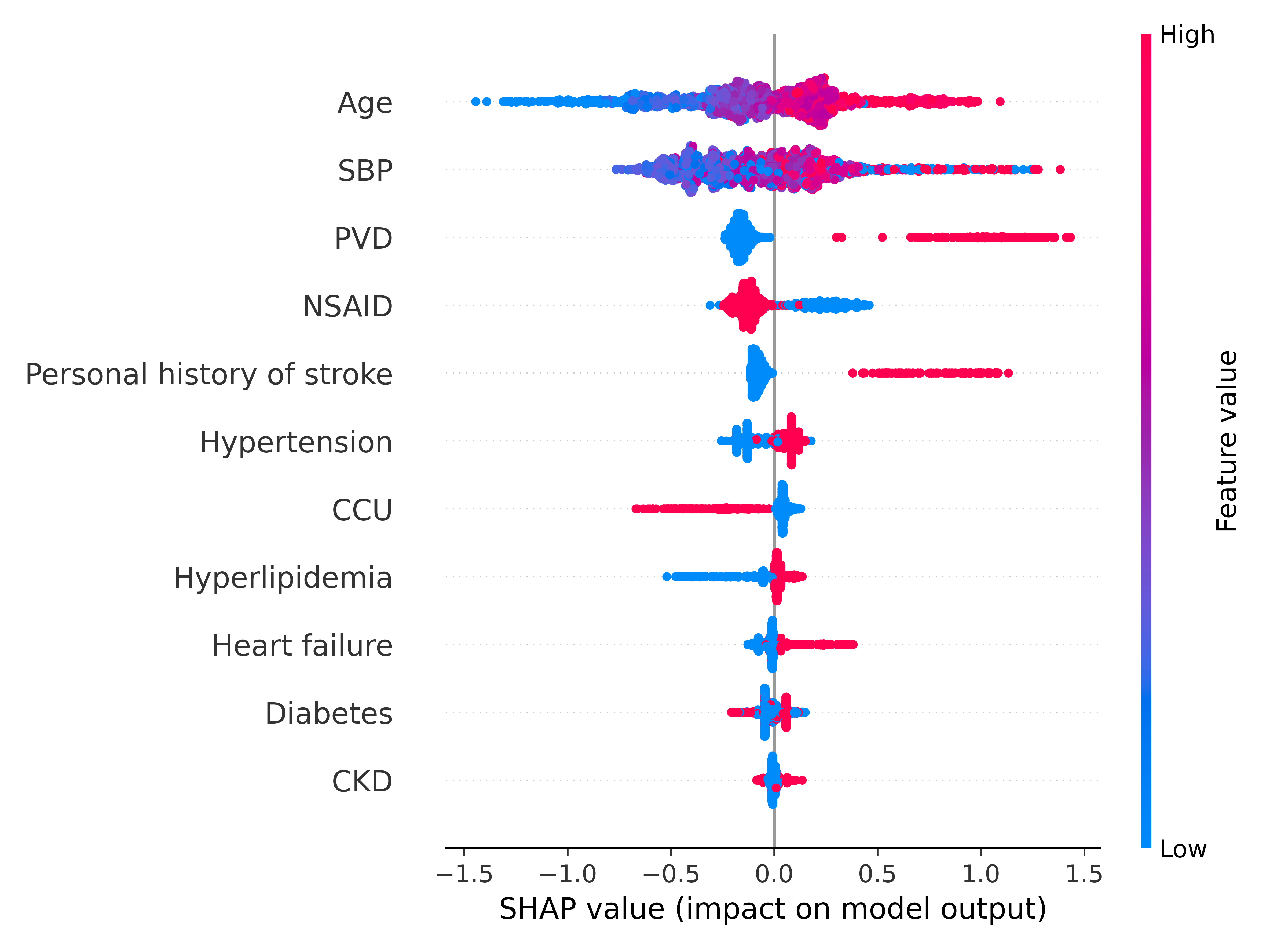}
    \caption{SHAP Beeswarm Plot (No CCI).}
    \label{fig7}
\end{figure}

Following the removal of CCI from the model (Fig~\ref{fig7}), the SHAP contributions of various variables exhibited notable shifts: the significance of age increased, whereas the contributions of CKD, heart failure, and diabetes decreased, with their SHAP values reversing direction. The implications of this effect and its potential influence on stroke prediction will be examined in the Discussion section.

\section*{Discussion}

\subsection*{Comparison with Previous Research}
The proposed model of this study, SVM, exhibited notable advantages in predictive performance (Table~\ref{table4}), especially regarding AUC, which attained a value of 0.855 (95\% CI: 0.829–0.878). In comparison, the CatBoost model developed by Lin et al \cite{bib7}. attained a mere 0.760 (95\% CI: 0.722–0.798), suggesting that our model exhibits a superior overall discriminative capability. While both models demonstrated a sensitivity of around 0.720, indicating comparable efficacy in accurately identifying high-risk patients, our SVM model attained a specificity of 0.817, in contrast to the comparison model's 0.660. The elevated specificity illustrates the model's enhanced capability in precisely identifying and excluding non-high-risk patients, thus minimizing misclassification and the inefficient use of resources. Furthermore, the comprehensive accuracy of our SVM model reached 0.809, significantly surpassing the 0.664 recorded for Lin et al.'s model.

Despite the optimized feature selection methods and advanced machine learning algorithms, another significant contributing element is our deliberate incorporation of specific comorbidities—such as diabetes, heart failure, and CKD—as distinct predictors, rather than depending exclusively on the CCI. Lin et al. \cite{bib7} illustrated that CCI was a predominant factor in their SHAP analysis; however, its application as a composite multimorbidity index might inadvertently obscure the distinct impacts of high-risk comorbidities.

\begin{table}[H]
\begin{adjustwidth}{-1.25in}{0in}
\caption{Model Performance Comparison with Previous Research}
\label{table4}
\begin{tabular}{l|l|c|c|c}
\hline
Model  & AUC (95\%CI) & Sensitivity & Specificity & Accuracy (95\%CI) \\
\hline
SVM                 & 0.855 (0.829 - 0.878) & 0.720 & 0.817 & 0.809 (0.792 - 0.826) \\ \hline
Lin et al. \cite{bib7}   & 0.760 (0.722-0.798) & 0.718 & 0.660 & 0.664 (0.642-0.687) \\
\hline
\end{tabular}
\end{adjustwidth}
\end{table}

A comparable finding was noted in research conducted by Park et al. (2018), which assessed predictive models concerning mortality among kidney transplant recipients. Their research indicated that the CCI alone exhibited inadequate discrimination in predicting patient survival\cite{bib24}. However, when specific comorbidities—such as diabetes, myocardial infarction, and PVD—were explicitly integrated into the model, there was a notable enhancement in predictive accuracy. This discovery underscores the notion that although CCI serves as a valuable tool for assessing overall comorbidity burden, the integration of particular high-risk conditions as independent variables can significantly improve model efficacy in disease-specific forecasts.

The analysis of SHAP importance (Fig~\ref{fig5}) further substantiates this differentiation, illustrating that CCI stands as the predominant predictor, while the explicit inclusion of CKD, diabetes, and heart failure significantly augments the model's predictive efficacy. This supports our hypothesis that depending just on composite indices like CCI could restrict the ability of the algorithm to identify subtle, disease-specific risk patterns. On the other hand, treating individual comorbidities as distinct features offers a more detailed and clinically relevant risk assessment, thereby enhancing the accuracy of stroke predictions. Consequently, although Lin et al. (2024) confirmed the predictive efficacy of CCI in machine learning-driven stroke prediction, our findings show that a hybrid approach—using both CCI and major comorbidities separately—improves predictive accuracy even more.

\subsection*{Reallocation of Feature Contributions in SHAP}
A notable finding in the SHAP study (Figure~\ref{fig6}) was the unexpected negative SHAP values for age, CKD, diabetes, and heart failure, which initially appeared counterintuitive given their established associations with increased stroke risk\cite{bib11,bib16}. This phenomena presumably results from CCI serving as a major predictor that integrates multimorbidity-related risk, therefore diminished the apparent independent contributions of these individual comorbidities \cite{bib25}.  

The computation of SHAP is based in the principles of Shapley value theory, and in instances where numerous variables display significant multicollinearity, SHAP may adjust the distribution of feature contributions accordingly \cite{bib26}. Li et al. (2020) in their study illustrate that when a predominant variable, like the Gleason score, has effectively encapsulated the majority of risk information pertinent to the disease, related covariates, such as prostate-specific antigen (PSA), may display negative SHAP values. However, this does not imply that these factors intrinsically reduce risk; rather, it indicates that the model emphasizes the collective data from the major predictor, hence diminishing the incremental effects of individual variables \cite{bib27}. This allies with our findings, wherein the CCI has effectively captured the risk burden associated with age, CKD, diabetes, and heart failure, thereby resulting in negative SHAP attributions for these variables. 

Furthermore, Jafari et al. (2022) conducted an investigation into the stratification of chronic disease risk and found that when body mass index (BMI) or metabolic syndrome had predominantly accounted for the majority of body composition-related risk, other highly correlated features, such as waist circumference and hip circumference, demonstrated negative SHAP contributions despite their recognized prognostic importance \cite{bib28}. This emphasized that negative SHAP values should not be misinterpreted as a protective effect but rather as an artifact of feature collinearity and the redistribution of model-derived attributions. In our study, CCI, as a composite comorbidity burden index, has already consolidated the predictive contributions of multiple risk factors \cite{bib8}. A direct comparison between Figure~\ref{fig6} and Figure~\ref{fig7} further reinforces this interpretation. Following the exclusion of CCI, a marked alteration in SHAP contributions was noted—most prominently, the significance of age rose considerably, whereas the contributions of CKD, diabetes, and heart failure diminished though their SHAP values adjusted positively in relation to stroke risk. This suggests that CCI had previously absorbed a substantial portion of the risk-related information associated with these comorbidities, leading to an artificial suppression of their individual SHAP contributions. As a result, the removal of CCI compelled the model to reassign risk to the individual variables, thereby reinstating their positive correlations with stroke risk. Nevertheless, the overall predictive significance of these factors was inferior to that observed when CCI was incorporated, suggesting that their roles were largely subsumed by CCI in the initial model.

\subsection*{Impact of SHAP Negative Values on Model Predictive Performance}
Despite the negative SHAP values observed for these variables, the model's AUC remained at 0.855 (Table~\ref{table3}), indicating that these features are still crucial to overall predictive performance. Similarly, the study by Li et al. (2020) found that even though PSA exhibited negative SHAP values, removing PSA from the model resulted in a decrease in AUC, demonstrating that it still contained valuable predictive information\cite{bib27}. In our study, we conducted an ablation analysis by removing CKD, diabetes, and heart failure, which led to a decline in AUC to 0.790. This further confirms that even though these variables may appear to have negative SHAP contributions, they remain essential for model prediction.

\subsection*{Limitation}
This study has achieved notable advancements in predicting postoperative stroke risk among patients with coronary artery disease (CAD) undergoing revascularization, utilizing refined feature selection methods and sophisticated machine learning techniques; however, some limitations persist.

Despite a thorough literature review and expert consultation throughout the feature selection process, the inclusion of comorbidities that are already part of the CCI as independent predictors unavoidably leads to multicollinearity challenges. Future research may explore potential solutions such as modifying the weighting scheme of CCI calculations or explicitly modeling the interrelationships among its component diseases. This approach could help alleviate multicollinearity effects while improving both the predictive performance and interpretability of the model.

Moreover, this study is deficient in external validation, as the evaluation of model performance was conducted exclusively within the confines of the MIMIC-IV database, without the corroboration of an external dataset. This constraint could hinder the model's applicability in various clinical environments or healthcare organizations. Future investigations ought to integrate independent, multicenter external validation datasets to thoroughly evaluate the model’s applicability and reliability in real-world scenarios, thus enhancing the robustness of its predictive outcomes and its clinical utility.

\section*{Conclusion}
This study, based on the MIMIC-IV database, targeted patients undergoing coronary revascularization procedures (PCI or CABG) and developed as well as evaluated multiple machine learning models for predicting postoperative stroke risk. With a test set AUC of 0.855, the SVM model showed the best predictive capability among all the models, outperforming conventional logistic regression and the previously reported CatBoost model. Moreover, the analysis of feature importance utilizing SHAP revealed that the CCI remained to be a major postoperative stroke risk predictor.  Additionally, including specific comorbidities—diabetes, CKD, and heart failure—as independent variables markedly improved the model's predictive efficacy. 

It is necessary to note that while the CCI played a crucial role in the overall risk assessment, it also "absorbed" a portion of the risk information from high-risk comorbidities, resulting in negative SHAP contribution values for certain established risk factors (e.g., diabetes, CKD, heart failure). This phenomenon highlights the strong correlation between the CCI and these comorbidities, with the model attributing the primary risk weight to the CCI, consequently reducing the marginal contributions of the other features. Upon the subsequent removal of the CCI, the contributions and directives of these features were meticulously recalibrated, thereby affirming their genuine significance in the assessment of stroke risk. 

In conclusion, the research demonstrates that incorporating both the CCI and specific comorbid conditions, such as diabetes, CKD, and heart failure, into the predictive model significantly strengthens the evaluation of stroke risk following coronary revascularization. This integrated methodology, unlike approaches that rely solely on CCI, takes into account the additional risks associated with specific comorbidities, thereby enhancing the overall predictive efficacy of the model.  The results present healthcare professionals with a more sophisticated instrument for preoperative evaluation and tailored intervention, which may enhance outcomes for patients at elevated risk and decrease the likelihood of postoperative stroke. Future endeavors should prioritize refining the CCI by re-weighting its components for more accurate stroke-risk estimation, as well as conducting external validation to confirm the model’s generalizability across diverse clinical settings.

\section*{Author Contributions}
\begin{description}
    \item[\bf Conceptualization:] Haonan Pan, Shuheng Chen, Maryam Pishgar
    \item[\bf Data curation:] Haonan Pan
    \item[\bf Formal analysis:] Haonan Pan
    \item[\bf Investigation:] Haonan Pan, Elham Pishgar, Kamiar Alaei, Greg Placencia
    \item[\bf Methodology:] Haonan Pan, Shuheng Chen
    \item[\bf Visualization:] Haonan Pan
    \item[\bf Writing – original draft:] Haonan Pan
    \item[\bf Writing – review \& editing:] Haonan Pan, Shuheng Chen, Maryam Pishgar
\end{description}


%
%
%

\end{document}